\documentclass[a4paper,10pt]{elsarticle}

\usepackage{lineno,hyperref}
\modulolinenumbers[5]
\usepackage[T1]{fontenc}
\usepackage[utf8]{inputenc}
\usepackage{textcomp}
\usepackage{gensymb}
\usepackage{url}            
\usepackage{booktabs}       
\usepackage{amsfonts}       
\usepackage{nicefrac}       
\usepackage{microtype}      
\usepackage{graphicx}
\usepackage{colortbl}
\usepackage{flushend}
\usepackage{multirow}
\usepackage{times}
\usepackage{natbib}
\usepackage{enumitem}
\usepackage[font=it]{caption}
\usepackage{subcaption}
\usepackage{doi}
\usepackage{hyperref}
\usepackage[usenames,dvipsnames]{xcolor}
\usepackage{listings}
\usepackage{array}
\usepackage{placeins}
\usepackage{psfrag}
\usepackage[top=2.5cm, bottom=2.5cm, left=1.5cm, right=1.5cm,
columnsep=0.8cm]{geometry}
\usepackage{setspace}

\definecolor{Gray}{gray}{0.9}

\usepackage{algorithmicx}
\usepackage{algorithm}
\usepackage{algpseudocode}
\usepackage{siunitx}
\usepackage{amsmath}
\usepackage{amsfonts}
\usepackage{amssymb}
\usepackage{mathtools}
\usepackage{eurosym, units}

\newcommand{\displaycomments}

\ifdefined\displaycomments

\usepackage{xtab}
\usepackage{verbatim}
\newlength\figureheight
\newlength\figurewidth
\usepackage{pgfplots}
\usepackage{epstopdf}
\usepackage{graphicx}
\usepackage{float}
\usepackage{color} 
\usepackage{circuitikz}

\begin{document}


\begin{frontmatter}

\title{Physics-constrained Deep Learning of Multi-zone Building Thermal Dynamics}
  
  \author[]{J\'an Drgo\v na\corref{a1}}
      \ead{jan.drgona@pnnl.gov}
  \author[]{Aaron R. Tuor}
        \ead{aaron.tuor@pnnl.gov}
  \author[]{Vikas Chandan}
        \ead{vikas.chandan@kpnnl.gov } 
  \author[]{Draguna L. Vrabie}
        \ead{draguna.vrabie@pnnl.gov } 

  \address{Pacific Northwest National Laboratory, Richland, WA, USA }
  \cortext[a1]{Corresponding author.}

  \begin{abstract}
We present a physics-constrained control-oriented deep learning method for modeling building thermal dynamics.
The proposed method is based on the systematic encoding of physics-based prior knowledge into a structured recurrent neural architecture. Specifically, our method incorporates structural priors from traditional physics-based building modeling into the neural network thermal dynamics model structure. Further, we leverage penalty methods to provide inequality constraints,  thereby bounding predictions within physically realistic and safe operating ranges. Observing that stable eigenvalues accurately characterize the dissipativeness of the system, we additionally use a constrained matrix parameterization based on the Perron-Frobenius theorem to bound the dominant eigenvalues of the building thermal model parameter matrices. We demonstrate the proposed data-driven modeling approach's effectiveness and physical interpretability on a dataset obtained from a real-world  office building with $20$ thermal zones.
Using only $10$ days' measurements for training, we demonstrate generalization over $20$ consecutive days, significantly improving the accuracy compared to prior state-of-the-art results reported in the literature. 
  \end{abstract}


\end{frontmatter}

\section{Introduction} \label{sec:intro}
 
Energy-efficient buildings are one of the top priorities to sustainably address the global energy demands and reduction of the CO$_2$ emissions~\cite{ML_climate2019,IEA2015}.
Advanced control strategies for buildings have been identified as a potential solution with projected energy saving potential up to $28\%$~\cite{HVAC_potential_2002,GyaEtal_2010_savings,DRGONA2020}.
The current state of the art approaches in the domain are based on constrained optimal control methods and heavily depend on the mathematical models of the building dynamics~\cite{Siroky11,ma2012model,maasoumy2013online,chandan2014decentralized,baldi2016dual,DRGONA202063}.

The main bottleneck in the deployment of model predictive control (MPC) in practice is the development of the building thermal dynamics model~\cite{cigler2013beyond}.
In general, the building thermal behavior is characterized by high-dimensional, nonlinear, and often discontinuous dynamics, for which modeling typically requires expertise and development time~\cite{IDEAS_verification,wet14,GAO2019}. 
Moreover, high computational demands and non-differentiability can easily cast the white-box model as not suitable for efficient gradient-based optimization that is typically used in various applications.
Therefore, successful control applications require
 a trade-off between model accuracy and computational efficiency.
Data-driven system identification and controller approximations typically represent more computationally efficient alternatives~\cite{ROMAN2020109972,ZHANG2015177,DPC_Smarra,Ferkl2010205,DRGONA2018}.
However, purely-black box models require a large amount of data and may not generalize well outside the training distribution~\cite{AFRAM2014_models,AFROZ2018}. On the other hand, identifying accurate and reliable gray-box models remains a challenging task and involves solving difficult non-convex optimization problems~\cite{Arroyo2020,ModestPy,MPCairport}.
 As a consequence, many of the current control-oriented modeling approaches for buildings still rely on crude approximations assuming low-order linear dynamics, which may hamper the overall control performance~\cite{PICARD20177}.

In contrast with classical white-, gray-, and black-box modeling approaches, we present a novel data-driven modeling paradigm for building thermal dynamics based on constrained deep learning.
 We show that we can greatly decrease the required modeling expertise by systematically encoding structural assumptions and constraints while achieving a state of the modeling performance. The proposed neural architecture represents a generic abstraction applicable to an arbitrary building without manual re-design.

\subsection{Related Work}

\paragraph{Control-oriented building thermal modeling}
 When properly calibrated, white-box building models can generate highly accurate and reliable results~\cite{IDEAS_verification}.
However, despite  significant engineering effort invested in
development, due to multiple sources of uncertainty effecting the non-trivial and time-intensive model parameter tuning,  white-box models can still result in inaccurate predictions~\cite{Arendt2018}.
  Moreover, integrating computationally heavy white-box models in real-time optimization routines represents a major challenge and requires significant investment in software development~\cite{Henze_commercial,TACO}.

Due to the higher complexity and cost of white-box models, there is a rising trend towards the use of data-driven methods for predictive control of buildings~\cite{RATZ2019109384,GUO2020110500}.
 However, most of the data-driven approaches in the literature
are overly simplified and mostly applied to single-zone buildings~\cite{KATHIRGAMANATHAN2021110120}.
The most common type of data-driven method is based on linear system identification of reduce order models~\cite{PRIVARA2013113,rey14,ZAKULA2014}.
Many classical system identification methods typically minimize only  $1$-step ahead prediction errors.
However, as pointed out in~\cite{PRIVARA2013113}, optimizing over
multi-step ahead prediction errors is much more suitable for predictive control applications. This method is often referred to as  MPC relevant identification (MRI)~\cite{GOPALUNI2004699,LAURI2010118}, which is based on linear autoregressive models solved via partial least squares (PLS) algorithm. 
Nonetheless, authors in~\cite{PICARD20177} demonstrated that low-order linear models may not always provide sufficient accuracy and can negatively influence overall MPC performance.
On the other hand, machine learning models based on
regression trees~\cite{DPC_Smarra,DPC_RT_Jain2017}, or
neural networks\cite{kusiak:neural:2012,Ruano06,Huang201586} have demonstrated the capability of capturing nonlinear relationships in the building dynamics. 
However, more complex data-driven models have a tendency to overfit on small datasets~\cite{AFRAM2014_models,AFROZ2018,Arendt2018}.
 The main drawback of the mentioned black-box models is that they do not incorporate prior knowledge and may provide unreliable predictions by violating underlying physical laws.

Probabilistic semi-physical modeling (PSPM) based on stochastic differential equations~\cite{ANDERSEN200013,BACHER20111511}
falls into the category of gray-box methods requiring prior expert knowledge and manual design of case-specific system dynamics. 
Resistance-capacitance (RC) networks represent arguably the most popular gray-box modeling method~\cite{ModestPy, GBT:DeConinck}.
However, estimating parameters of complex RC networks for large-scale buildings leads to difficult to solve non-convex optimization problems, often requiring manual or heuristic tuning, further increasing the development time and cost~\cite{ModestPy,Arroyo2020}. 
 For further discussions and comparisons on different modeling methods used in building control, we refer the reader to the reviews in~\cite{ROMAN2020109972,AFROZ2018,boodi2018}.

\paragraph{Physics-constrained deep learning}
Incorporating constraints into neural networks has proven challenging due to non-convexity, a convergence of the learning process, and a requirement for rigorous guarantees of constraint satisfaction~\cite{ConstrainedML2019}.
Do to these issues, loss function augmentation via regularization and penalty methods has become the most popular way of imposing constraints in deep learning~\cite{PathakKD15,ConstrCNN7971941,conOrdinalReg2018}.
Although sacrificing strict bounds, it has been shown that soft constrained penalty methods work well in practice and often outperform methods based on hard constraints, such as barrier methods~\cite{MarquezNeilaSF17, logbarrierCNN2019}.
Alternatively, architecture design methods focus on incorporating strong inductive biases inspired by physics, for instance, architectures respecting energy conservation laws.
Examples of such architecture are linear operator constraints~\cite{hendriks2020linearly},
Hamiltonian~\cite{HamiltonianDNN2019} or Lagrangian~\cite{LagrancianDNN2019} neural networks.
From an architecture perspective, the work we present here is inspired by a family of neural state-space models (SSM)~\cite{krishnan2016structured,LatentDynamics2018,MastiCDC2018,NIPS2018_8004,OgunmoluGJG16}, representing structurally modified vanilla RNNs tailored for the modeling of dynamical systems for control. 

In recent years several authors interpreted deep neural networks through the optics of differential equations~\cite{NIPS2018_7892_NeuralODEs,DeepXDE2019,HeZRS15}. 
This new line of research opened the doors for a more rigorous analysis of the neural dynamics. For instance, authors in~\cite{HaberR17} linked
the vanishing and exploding gradient problems in recurrent neural networks (RNN) with the eigenvalues and stability of neural networks. Others proposed new architectures with stability guarantees based on constraining the eigenvalues of neural network layers~\cite{IMEXnet2019,NIPS2019_9292}.
In this paper, we leverage the eigenvalue constraints method based on Perron-Frobenius theorem~\cite{tuor2020constrained} interpreted through the perspective of building physics.
Additionally, we use penalty methods to constrain the learned  dynamics' phase space within physically realistic bounds.

\subsection{Contributions}
This paper shows how to train physics-constrained recurrent neural dynamics models tailored to efficiently learn the building's thermal dynamics in an end-to-end fashion, with physically coherent generalization, from small datasets.
This work is a conceptual extension of 
author's previous work on using constrained-deep learning models for system identification and control of small-scale building thermal dynamics model~\cite{tuor2020constrained,Drgona_stable}.
We empirically demonstrate the accuracy and generalization of the proposed physics-constrained neural architectures using only $10$ days of training data from a real-world office building with $20$ thermal zones.
 Compared to their unstructured and unconstrained counterparts, the presented constrained recurrent neural models show a $15\%$ reduction in error. 
We introduce several novel key features of the presented modeling approach resulting in increased data-efficiency, accuracy, generalization, and systematic constraint handling compared to unconstrained deep learning models:
\begin{enumerate}
    \item Encoding the underlying graph structure of the building dynamics as block-structured recurrent neural dynamics models.
    \item Eigenvalue constraints based on Perron-Frobenius theorem yealding guarantees on stability and dissipativity of  learned dynamics.
    \item Penalty methods for imposing inequality constraints representing physically meaningful boundary conditions of  learned dynamics. 
    \item Multi-step multi-term loss function for learning long-term and physically coherent dynamical models.
    \item Physical interpretation of eigenvalue analysis of neural network weights for inspection of the desired dynamical properties of the learned models.
\end{enumerate}

To the author's best knowledge, this is the first combined use of structured recurrent neural architectures with physics-inspired constraints applied to a real-world building thermal dynamics modeling problem.

\section{Methods}

\subsection{Building Thermal Dynamics}
\begin{figure*}[!htbp]
\centering
 \begin{subfigure}[b]{0.49\textwidth}
         \centering
        \includegraphics[width=0.9 \textwidth]{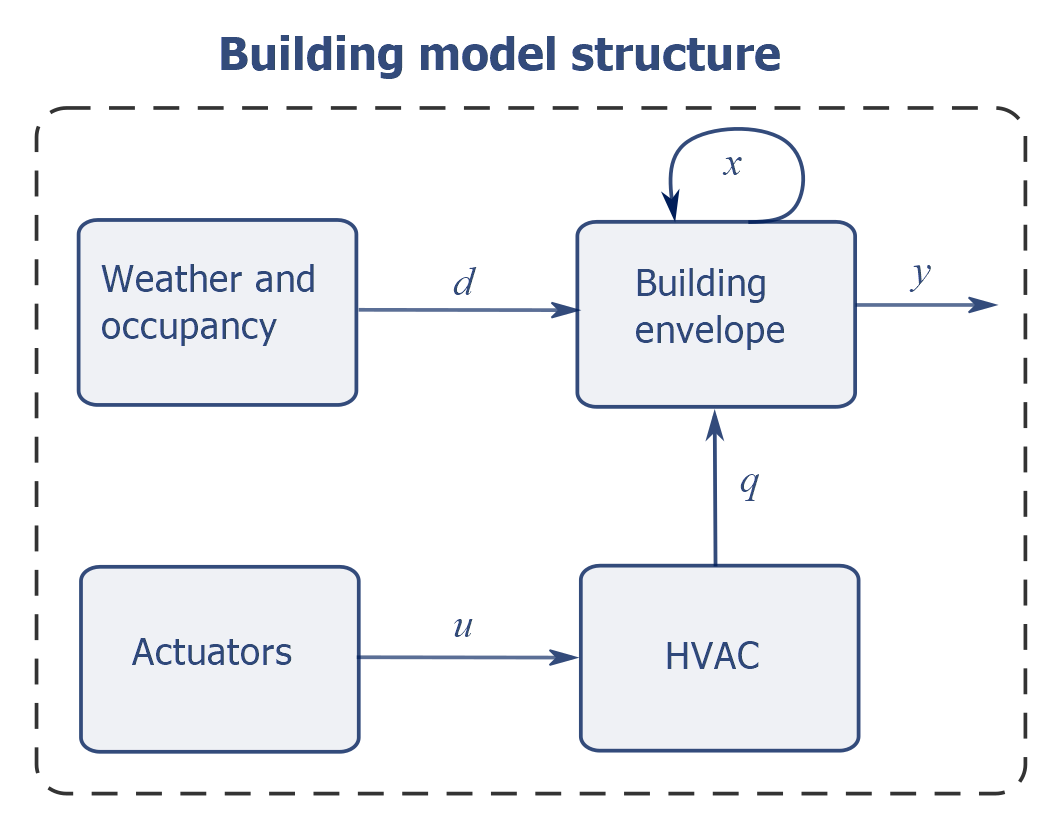}
         \caption{Structure of physics-based building thermal model.}
         \label{fig:building}
     \end{subfigure}
     \hfill
     \begin{subfigure}[b]{0.49\textwidth}
         \centering
         \includegraphics[width=0.85 \textwidth]{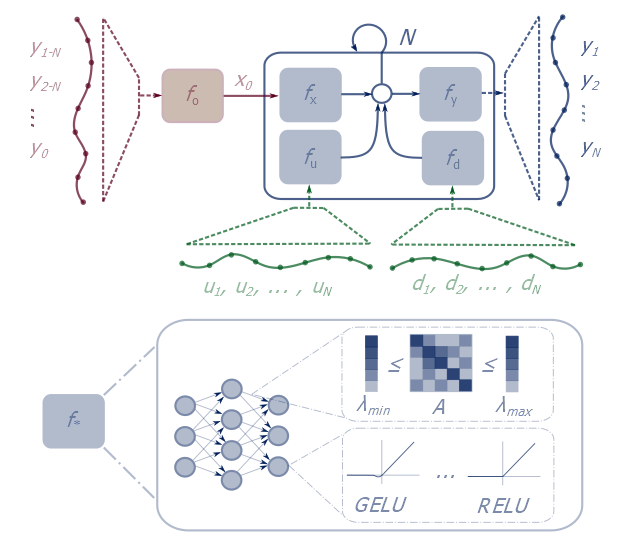}
         \caption{Structured recurrent neural dynamics model.}
         \label{fig:model}
     \end{subfigure}
\caption{Generic structure of physics-inspired recurrent neural dynamics model architecture. Weights of individual neural blocks $f_*$ are parametrized by linear maps with  constrained eigenvalues, while component outputs are subject to penalty constraints parametrized by common activation functions.}
\label{fig:building_model}
\end{figure*}

When developing predictive models for control purposes, one has to balance model complexity, robustness, and accuracy.
The typical building envelope dynamics is represented by a model with a graph structure shown in Fig.~\ref{fig:building}.
Mathematically, the thermal building model is given as the following difference equation with nonlinear input and disturbance dynamics:
\begin{subequations}
\label{eq:building:model}
\begin{align}
   \mathbf{x}_{t+1}  & = A \mathbf{x}_{t} + B \mathbf{q}_{t}  + f_d(\mathbf{d}_{t}),   \\
    \mathbf{y}_{t} & = C \mathbf{x}_{t},  \\
    \mathbf{q}_{t}  & = \dot{\mathbf{m}}_{t}  cp \Delta \mathbf{T}_{t},  \label{eq:building:model:hf}
\end{align}
\end{subequations}
where $\mathbf{x}_{t}$ and $\mathbf{y}_{t}$ represent the values of the states (envelope temperatures), and measurements (zone temperatures) at time $t$, respectively. 
Disturbances $\mathbf{d}_{t}$ represent the influence of weather and occupancy behavior.
Heat flows delivered to the building $\mathbf{q}_{t} $
are represented by~\eqref{eq:building:model:hf} heat flow equation as a product of mass flows $\dot{\mathbf{m}}_{t}$, difference of the supply and return temperatures $\Delta \mathbf{T}_{t}$, 
and the specific heat capacity constant $cp$.
Tab.~\ref{tab:variables} summarizes the variables used for data-driven modeling of the underlying building thermal dynamics.
\begin{table}[!htb]
	\captionof{table}{Abstract variable notation of the data-driven neural model of the building thermal dynamics.}
	\label{tab:variables}
  \centering
    \begin{tabular}{crrr}
    \toprule
    {Notation} &     
    {Neural model}  &  {Building physics}  & {Units}  \\
    \midrule
    $\mathbf{x}$  & hidden states & envelope temperatures,  heat flows  & [\SI{}{\kelvin}, \SI{}{\watt}] \\
    $\mathbf{y}$  & outputs & room operative temperatures  & [\SI{}{\kelvin}] \\
    $\mathbf{u}$  &  control actions &  mass flows, supply temperatures  & [\SI{}{\kilogram/\second}, \SI{}{\kelvin}] \\
     $\mathbf{d}$  & disturbances  &  ambient temperature & [\SI{}{\kelvin}] \\
     $\mathbf{s}$   &  slack variables & violations of boundary conditions  & [\SI{}{\kelvin}, \SI{}{\watt}]  \\
    \bottomrule
    \end{tabular}%
\end{table}

When the model is built with perfect knowledge from first principles, it is physically interpretable. For instance, the $A$ matrix represents 1-D heat transfer between the spatially discretized system states. $B$ matrix defines the temperature increments caused by the convective heat flow~\eqref{eq:building:model:hf} generated by the HVAC system, while $f_d$ captures highly nonlinear thermal dynamics  caused by the weather conditions or internal heat gains generated by the occupancy. 
However, every building represents a unique system with different operational conditions. Therefore, obtaining the parameters of the difference equations~\eqref{eq:building:model} from first principles is a time-consuming, impractical task.

\subsection{Unstructured Recurrent Neural Dynamics Model}
The white-box building model~\eqref{eq:building:model} is represented by a partially observable discrete ordinary differential equation (ODE) describing the evolution of the system's state over time.
Assuming zero prior knowledge about the model structure can translate this discrete ODE into an unstructured state-space model (SSM):
\begin{subequations}
\label{eq:black_box}
\begin{align}
\mathbf{x}_{t+1} &= f(\mathbf{x}_t, \mathbf{u}_t, \mathbf{d}_t) \\
\mathbf{y}_t &= f_{y}(\mathbf{x}_t) \\
\mathbf{x}_{0} &=  {f}_o([\mathbf{y}_{1-N}; \ldots; \mathbf{y}_{0}]) 
\end{align}
\end{subequations}
Where $f$ represents the main system dynamics, while $f_{y}$ defines the output mapping. 
We are using neural networks to encode the mappings
$f$ and $f_{y}$ and we refer to this model as 
unstructured recurrent neural dynamical model.

\subsection{Structured Recurrent Neural Dynamics Model}
The systems such as~\eqref{eq:building:model} with linear state dynamics and nonlinear input dynamics are formally referred to as block-structured state-space models. 
Conveniently, we can assume this generic structure to be invariant across building types and sizes.
Fig.~\ref{fig:model} shows the overall architecture of the physics-inspired neural dynamics model for partially observable systems. To further promote physically coherent behavior, the 
neural component blocks $f_*$ are parametrized by linear maps with constrained eigenvalues and regularized with penalty functions, as explained in the following sections.
The block-structured recurrent neural dynamics model is defined as:
\begin{subequations}
\label{eq:RNN}
\begin{align}
\mathbf{x}_{t+1} &= f_{x}(\mathbf{x}_t) + f_{u}(\mathbf{u}_t) + f_{d}(\mathbf{d}_t) \\
\mathbf{y}_t &= f_{y}(\mathbf{x}_t) \\
\mathbf{x}_{0} &=  {f}_o([\mathbf{y}_{1-N}; \ldots; \mathbf{y}_{0}]) 
\end{align}
\end{subequations}
Here $f_{x}$, $f_{u}$, and $f_{d}$ represent decoupled neural components of the overall system model, corresponding to state, input, and disturbance dynamics, respectively.
We assume only partially observable systems
where states $\mathbf{x}$ represent latent dynamics. As a consequence, we need to use state observer given as additional neural component, $f_{o}$, encoding a past $N$-step window of observations $\mathbf{y}$ onto initial state conditions $\mathbf{x}_0$.
During training, the model is unrolled and trained on an $N$-step ahead prediction window.
The main advantage of the block nonlinear over unstructured black-box state-space model lies in its structure. The decoupling allows us
to leverage prior knowledge for imposing structural assumptions and constraints onto individual blocks of the model. 
Please note the structural similarity between the proposed neural model~\eqref{eq:RNN} and difference equation~\eqref{eq:building:model}.

\subsection{Eigenvalue Constraints}
An important physics insight is that building thermal dynamics represents a  dissipative system with stable eigenvalues. 
The system's dissipativity is physically interpreted as heat losses of the building envelope, which are influenced by numerous factors such as building topology, material properties, insulation levels, or window-to-wall ratio.
These parameters determine the overall heat transfer coefficient
of the building,  called U-values, where smaller U-values mean better insulation.
The problem is that obtaining accurate information about the parameters required for estimation of the U-values  
from technical sheets is time consuming and tedious task.
From a dynamical perspective, U-values can be loosely related to the system eigenvalues. 
This inspired us to enforce physically reasonable constraints on the eigenvalues of a model's weight matrices. 

We leverage the method based on the Perron-Frobenius theorem, which states that the row-wise minimum and maximum of any positive square matrix defines its dominant eigenvalue's lower and upper bound, respectively. Guided by this theorem, we can construct a state transition matrix $\mathbf{\tilde{A}}$ with bounded eigenvalues:
\begin{subequations}
\label{eq:pf}
\begin{align}
\mathbf{M} &= \lambda_{\text{max}} - (\lambda_{\text{max}} - \lambda_{\text{min}}) \sigma(\mathbf{M'}) \\
\mathbf{\tilde{A}}_{i,j} &= \frac{\text{exp}(\mathbf{A'}_{ij})}{\sum_{k=1}^{n_x} \text{exp}(\mathbf{A'}_{ik})}\mathbf{M}_{i,j}
\end{align}
\end{subequations}
We introduce a matrix $\mathbf{M}$ which models damping parameterized by the matrix $\mathbf{M'} \in \mathbb{R}^{n_x \times n_x}$. 
We apply a row-wise softmax to another parameter matrix $\mathbf{A'} \in \mathbb{R}^{n_x \times n_x}$, 
then elementwise multiply by $\mathbf{M}$ to obtain our state transition matrix $\mathbf{\tilde{A}}$ with eigenvalues lower and upper bounds $\lambda_{\text{min}}$ and $\lambda_{\text{max}}$.
Further in the text we refer to this factorization as \texttt{pf} weight.  In case study, we compare the \texttt{pf} factorization with standard unconstrained 
 version referred to as \texttt{linear} weight.


\subsection{Inequality Constraints via Penalty Methods}
Using an optimization strategy known as the penalty method, 
we can add further constraints to our model such that its variables remain within physically realistic bounds. 
We enforce this property by applying inequality constraints via penalty functions $p(\mathbf{y})$ for each time step $t$:
\begin{subequations}
\label{eq:penalty}
\begin{align}
p(\mathbf{y}_t, \mathbf{\overline{y}}_t): \ \mathbf{y}_t - \mathbf{s}^{\overline{y}}_t \leq \mathbf{\overline{y}}_t  \:\:\: &\cong \:\:
    \mathbf{s}^{\overline{y}}_{t} = \text{max}(0,\:\mathbf{y}_t - \mathbf{\overline{y}}_t) \label{eq:penalty:ub} \\
p(\mathbf{y}_t, \mathbf{\underline{y}}_t): \ \mathbf{\underline{y}}_t \leq \mathbf{y}_t + \mathbf{s}^{\underline{y}}_t \:\:\: &\cong \:\:
    \mathbf{s}^{\underline{y}}_{t} = \text{max}(0,\:-\mathbf{y}_t + \mathbf{\underline{y}}_t) \label{eq:penalty:lb} 
\end{align}
\end{subequations}
 The constraints lower and upper bounds are given as $\mathbf{\underline{y}}_k$ and $\mathbf{\overline{y}}_k$, respectively. The slack variables $\mathbf{s}^{\underline{y}}_k$ and $\mathbf{s}^{\overline{y}}_k$ indicate the magnitude to which each constraint is violated, and we penalize them heavily in the optimization objective by a large weight on these additional terms in the loss function.
These constraints can be straightforwardly implemented using standard \texttt{RELU} functions (see right-hand sides of~\eqref{eq:penalty}) and included as auxiliary weighted terms in the loss function.

To understand how the penalties enforce the output value constraints, let's focus on the upper-bound penalty~\eqref{eq:penalty:ub}. Notice that if the difference between the output $\mathbf{y}_t$ and it's upper bound $\overline{y}_{t}$ is positive, then the output value is exceeding the bound, resulting in a non-zero value of the slack variable $\mathbf{s}^{\overline{y}}_{t}$. Otherwise, if the result is negative, the output must be within bounds, and the constraint is satisfied indicated by the zero-valued slack variable $\mathbf{s}^{\overline{y}}_{t}$. The same intuition can be applied to the lower-bound constraint~\eqref{eq:penalty:lb}.
Please note that penalty-based inequality constraints can be imposed on arbitrary model variables based on their assumed or known physical limits.

\subsection{Multi-step Multi-term Loss Function}
We optimize the following loss function augmented with regularization and penalty terms to train the recurrent neural model~\eqref{eq:RNN} unrolled over $N$ steps:
\begin{equation}
\label{eq:loss}
\begin{split}
\mathcal{L}_{\text{MSE}}(\mathcal{Y}^{\text{ref}}, \mathcal{Y} | \Theta) = \frac{1}{N} \sum_{t=1}^{N} 
||\mathbf{y}^{\text{ref}}_{t} - \mathbf{y}_{t}||^2_2 
 + Q_{\text{dx}}||\mathbf{x}_{t} - \mathbf{x}_{t-1}||^2_2 + \\ Q_{\text{ineq}}^{\mathbf{y}}||\mathbf{s}^{\mathbf{y}}_{t}||^2_2  +  Q_{\text{ineq}}^{\mathbf{u}}||\mathbf{s}^{f_u}_{t}||^2_2  + Q_{\text{ineq}}^{\mathbf{d}}||\mathbf{s}^{f_d}_{t}||^2_2 
 \end{split}
\end{equation}
The first term of the loss function computes the mean squared error between predicted $ \mathbf{y}$ and observed outputs $\mathbf{y}^{\text{ref}}$ over $N$ time steps and represents our primary objective.
Following similar arguments as in the case of MRI method~\cite{PRIVARA2013113}, optimizing over $N$-step prediction window improves the overall accuracy and generalization 
of the learned system dynamics model.
The term $\mathbf{x}_{t} - \mathbf{x}_{t-1}$ represents state difference penalty promoting learning of smoother 
and physically more plausible state trajectories.
The violations of the inequality constraints defining the boundary conditions of outputs $\mathbf{y}$, 
are penalized by  incorporating weighted slack variables $\mathbf{s}^{\mathbf{y}}$. 
Thanks to the block-structured dynamics, we can constrain the dynamical contribution of inputs $f_u$ and disturbances $f_d$ towards the overall dynamics via two additional terms in the loss function.
This allows us to limit the effect of the external factors to be bounded within physically plausible ranges.
For instance, it is not physically realistic that $1$ K change in the ambient temperature would cause a $2$ K change in indoor temperature in a single time step.

\section{Experimental Case Study}
 The objective is to  develop a control-oriented  model of the thermal dynamics of a commercial office building, given only a limited amount of time series measurement data.

\subsection{Real-world Building Dataset and Experimental Setup}

\paragraph{Real-world Building Dataset}
The building used in this study is a commercial building in Richland, WA described in \cite{rubio2017learning}. Heating and cooling are provided by a variable air volume (VAV) system served by 4 air handling units (AHUs) serving 24 VAV boxes (zones). Each VAV box is equipped with a hot water reheat coil. A boiler, fed by natural gas, supplies hot water to the reheat coils and AHU coils. Chilled water is supplied by a central chiller plant. 
\begin{figure*}[!htbp]
\centering
     \begin{subfigure}[b]{0.95\textwidth}
         \centering
         \includegraphics[width=1. \textwidth]{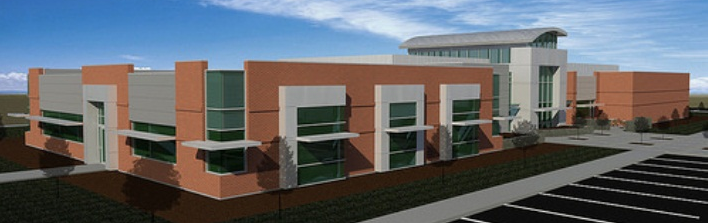}
         \caption{Building's facade.}
         \label{fig:building_real:facade}
     \end{subfigure}
      \begin{subfigure}[b]{0.95\textwidth}
         \centering
        \includegraphics[width=1.0 \textwidth]{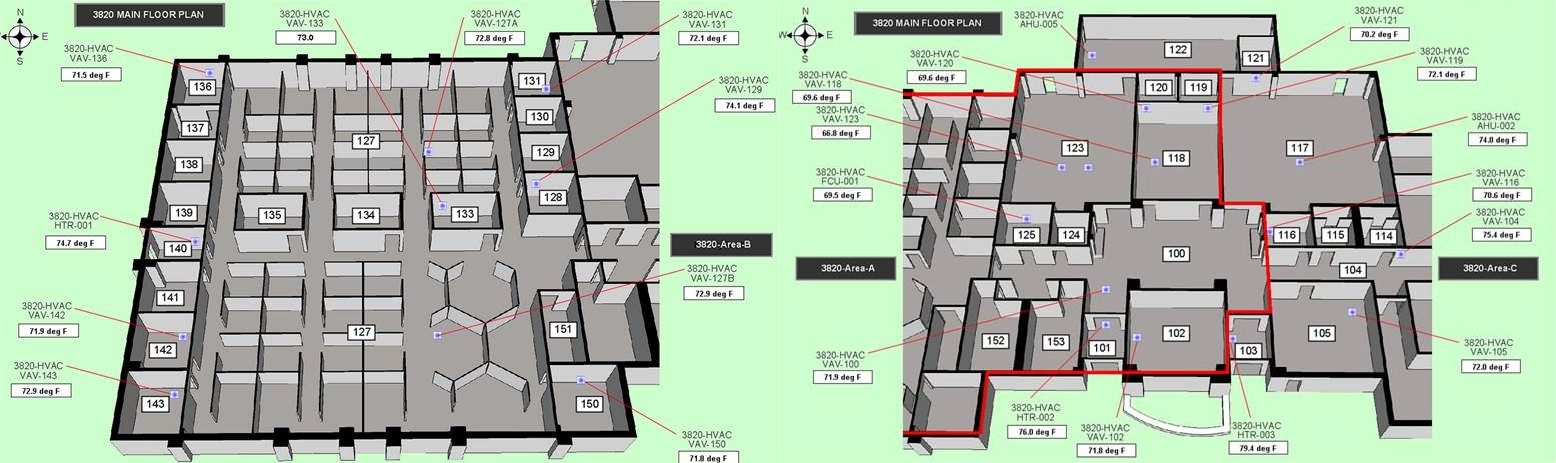}
         \caption{Building's zone layout.}
         \label{fig:building_real:layout}
     \end{subfigure}
\caption{Office building of interest.}
\label{fig:building_real}
\end{figure*}

Data from specific sensors for the above-mentioned buildings is stored in a database, which communicates with the building management system (BMS) and polls data for these sensors at a time resolution of 1 minute. A total of more than 600 sensors report data corresponding to measurements such as supply and return temperatures of air and water, air, hot water and cold water flow rates, energy and power consumption, set-points for the underlying control systems, occupancy status in zones, and outside air temperature. Data was cleaned and pre-processed according to the methodology described in \cite{rubio2017learning}.
The same dataset was used in~\cite{rubio2017learning} to model the building's power consumption and zone temperatures using RNN model with LSTM architecture. The authors have been able to achieve high prediction accuracy on a single step ahead prediction compared to other standard machine learning models such as linear regression, support vector regression, and random forests.
However, due to the purely black-box nature and $1$-step ahead loss function, the model in~\cite{rubio2017learning} does not explicitly guarantee physical constraints and is not suitable for long-term predictions of the building's thermal behavior. 
This case study demonstrates improved accuracy, generalization, long-term prediction capabilities, and physically coherent and interpretable dynamic behavior of the learned dynamical model with $20$ thermal zones. Hence, considering a model with higher complexity compared to the 2-zone model presented in~\cite{rubio2017learning}.

The time series dataset $D$ with datapoints relevant for system identification of thermal dynamics is given in the form of tuples with input, disturbance, and output variables, respectively.
\begin{equation}
    \label{eq:dataset}
    D = \{(\mathbf{u}^{(i)}_t, \mathbf{d}^{(i)}_t, \mathbf{y}^{(i)}_t), (\mathbf{u}^{(i)}_{t+\Delta}, \mathbf{d}^{(i)}_{t+\Delta}, \mathbf{y}^{(i)}_{t+\Delta}),  \ldots, (\mathbf{u}^{(i)}_{t+N\Delta}, \mathbf{d}^{(i)}_{t+N\Delta}, \mathbf{y}^{(i)}_{t+N\Delta})\},
\end{equation}
where $i = \mathbb{N}_1^n$ represents index of $n$ different batches of time series trajectories with $N$-step time horizon length. 
The data is sampled with fixed sampling time $\Delta = 15$ min.
We have in total $n_y = 20$ output variables corresponding to zone temperatures, 
$n_u = 40$ input variables representing HVAC temperatures and mass flows, and $n_d = 1$ disturbance variable representing ambient temperature forecast.
 Before training, we use \texttt{min-max} normalization to scale all variables between $[0, 1]$.
The dataset consists of $30$ days, which corresponds to only $2880$ datapoints.
We group the dataset into evenly split training, validation, and test sets, $960$ data points each.
We select the best performing models on the open-loop MSE for the development set and report results on the test set.


\paragraph{Experimental Setup}
We implement the presented model architectures using Pytorch~\cite{paszke2019pytorch}, and train with randomly initialized weights using the Adam optimizer \cite{kingma2014adam} with a learning rate of $0.003$, and $5,000$ gradient descent updates. We select the best performing model on the development set from a directed hyperparameter search. 
All neural network blocks are designed with \texttt{GELU} activation functions~\cite{HendrycksG16}.
The state estimator is encoded with a fully connected neural network, while individual neural blocks $f_*$ are represented either  by standard multilayer perceptron (MLP), recurrent neural network (RNN), or residual neural network (ResNet), respectively, each with $2$ layers and $80$ nodes.
We range the prediction horizon as powers of two $2^n$ with $n = 3, \ldots, 6$, which corresponds to $2$ up to $16$  hour prediction window.
The relative weights of the multi-term loss function for constrained models are $Q_{\text{dx}}=0.2$, $Q_{\text{ineq}}^{\mathbf{y}}=1.0$, $Q_{\text{ineq}}^{\mathbf{u}}=0.2$, and $Q_{\text{ineq}}^{\mathbf{d}}=0.2$. We set $\lambda_{\text{min}} =0.8$ and $\lambda_{\text{max}}=1.0$ for stability and low dissipativity of learned dynamics when using eigenvalue constraints.

\subsection{Results and Analysis}
This section assesses the open-loop and $ N $-step simulation performance of trained recurrent neural dynamics models with and without structure and constraints, respectively.
We systematically compare and analyze the added value of the block structure, penalty, and eigenvalue constraints, where 
Tab.~\ref{tab:best_results} summarizes the best performance of the modeling variants. 
Moreover, we discuss the interpretability of the proposed data-driven models through the optics of building physics.
\begin{table}[!htbp]
\caption{Test set MSE of best-performing structured, unstructured, constrained and unconstrained models, respectively.} 
        \centering
        \begin{tabular}{lcccrr}
        \toprule
           Structure       & Constrained   & Weights &   $N$  & $N$-step [K]     & Open-loop  [K] \\ \midrule
            \multirow{2}{*}{Structured}   & Y  & Linear  & 64   &  0.4811 &  0.4884 \\
              & N  & Perron-Frobenius   & 16   &  0.4720 &   0.5043 \\
            \multirow{2}{*}{Unstructured} & Y & Linear  & 64
             & 0.5380 & 0.5446 \\
            & N & Linear & 16
             & 0.5266 & 0.5596 \\
        \bottomrule
        \end{tabular}
        \label{tab:best_results}
    \end{table}

\begin{figure}[!htbp]
\centering
\includegraphics[width=1.0 \textwidth]{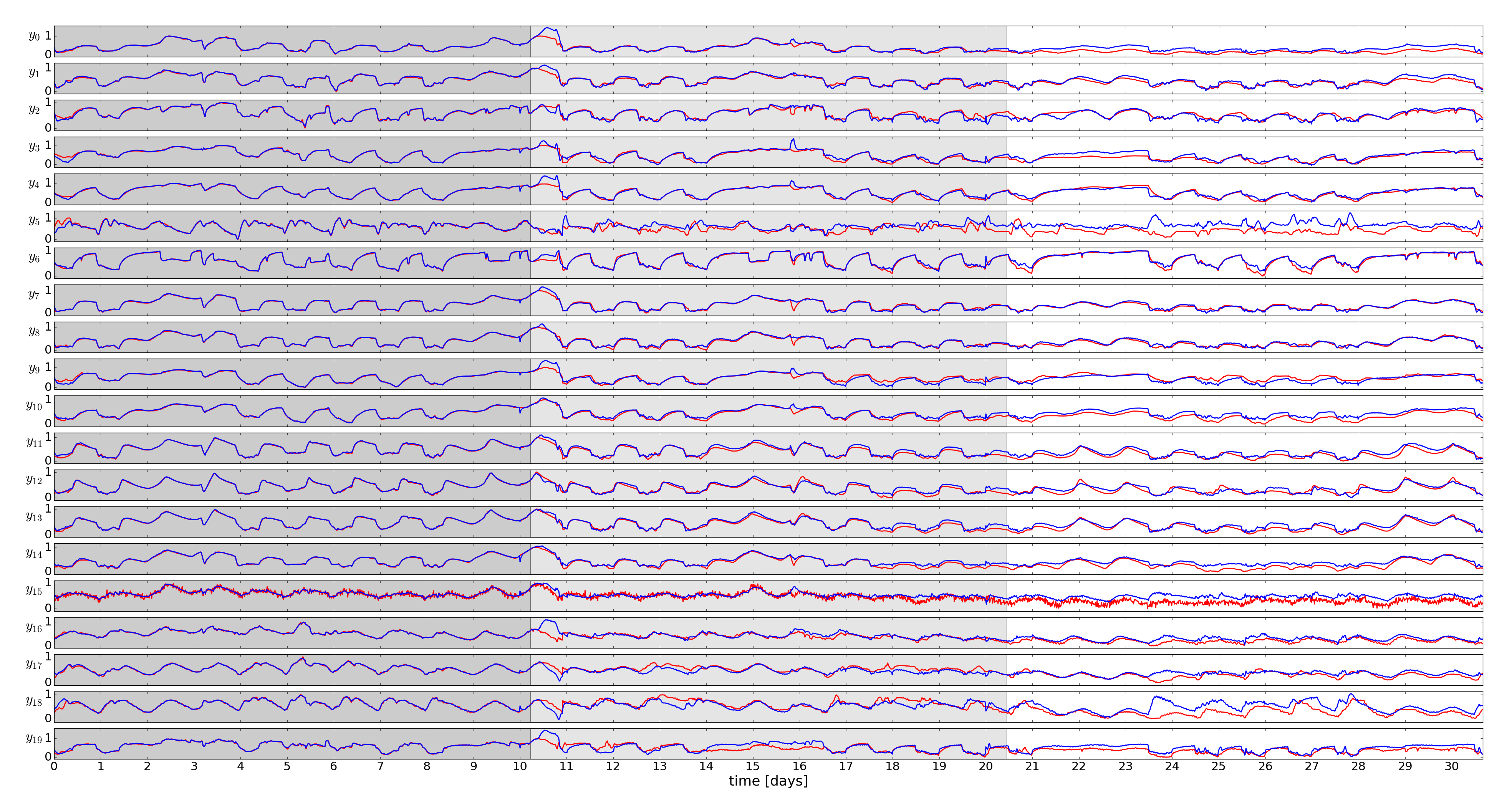}
\caption{Open-loop trajectories of the learned (blue) and ground truth (red) multi-zone building thermal dynamics.}
\label{fig:open_loop}
\end{figure}

\paragraph{Best performing model} 
As reported in Tab.~\ref{tab:best_results}, we achieve the best performance with constrained and structured recurrent neural model~\eqref{eq:RNN}.
The best model scores $0.0052$, $0.0091$, and $0.0143$, on normalized open-loop MSE evaluated on the test, dev, and train set, respectively.
From a physical perspective, the denormalized open-loop MSE corresponds to roughly $0.18$K, $0.31$K, and $0.49$K errors per output, respectively.
This demonstrates the ability to generalize the dynamics over the period of $30$ days, given only $10$ days of training data.
In comparison, the state of the art gray-box and black-box system identification methods trained on a similar amount of data reports open-loop MSE greater than $1.0$K~\cite{Arroyo2020,Mugnini13123125,picard2016comparison}. 
Hence our results show more than $100\%$ improvement against state of the art. 
However, a more rigorous comparison needs to be performed to compare the accuracy with standard gray-box methods using the same datasets.
For visual assessment, Fig.~\ref{fig:open_loop} shows normalized open-loop simulation trajectories of best performing structured dynamics model on the train, dev, and test set, represented by gray zones, respectively.

\begin{figure*}[!htbp]
\centering
 \begin{subfigure}[b]{0.49\textwidth}
         \centering
        \includegraphics[width=1.0 \textwidth]{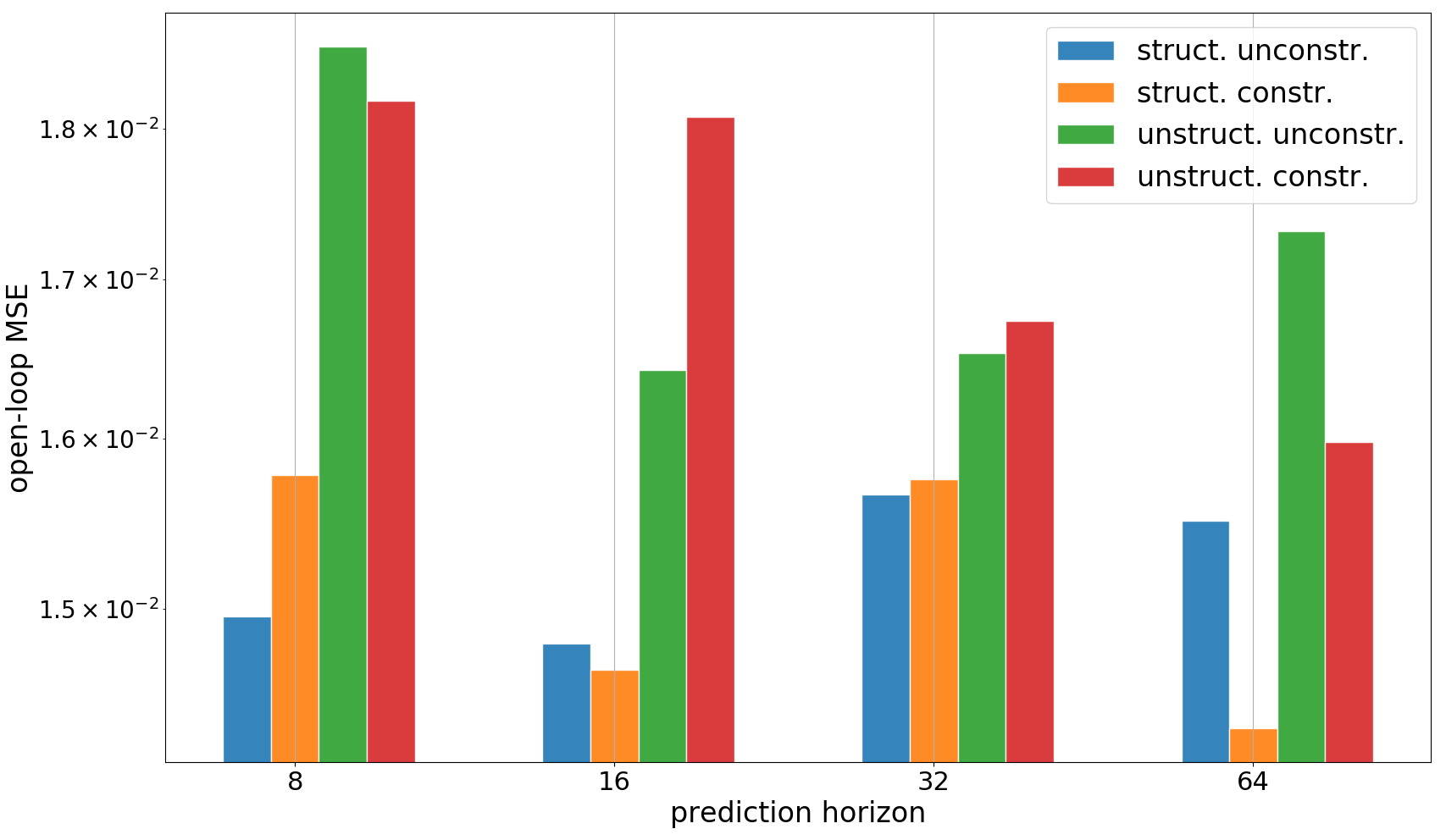}
         \caption{Open-loop MSE.}
         \label{fig:openMSE}
     \end{subfigure}
     \hfill
     \begin{subfigure}[b]{0.49\textwidth}
         \centering
         \includegraphics[width=1.0 \textwidth]{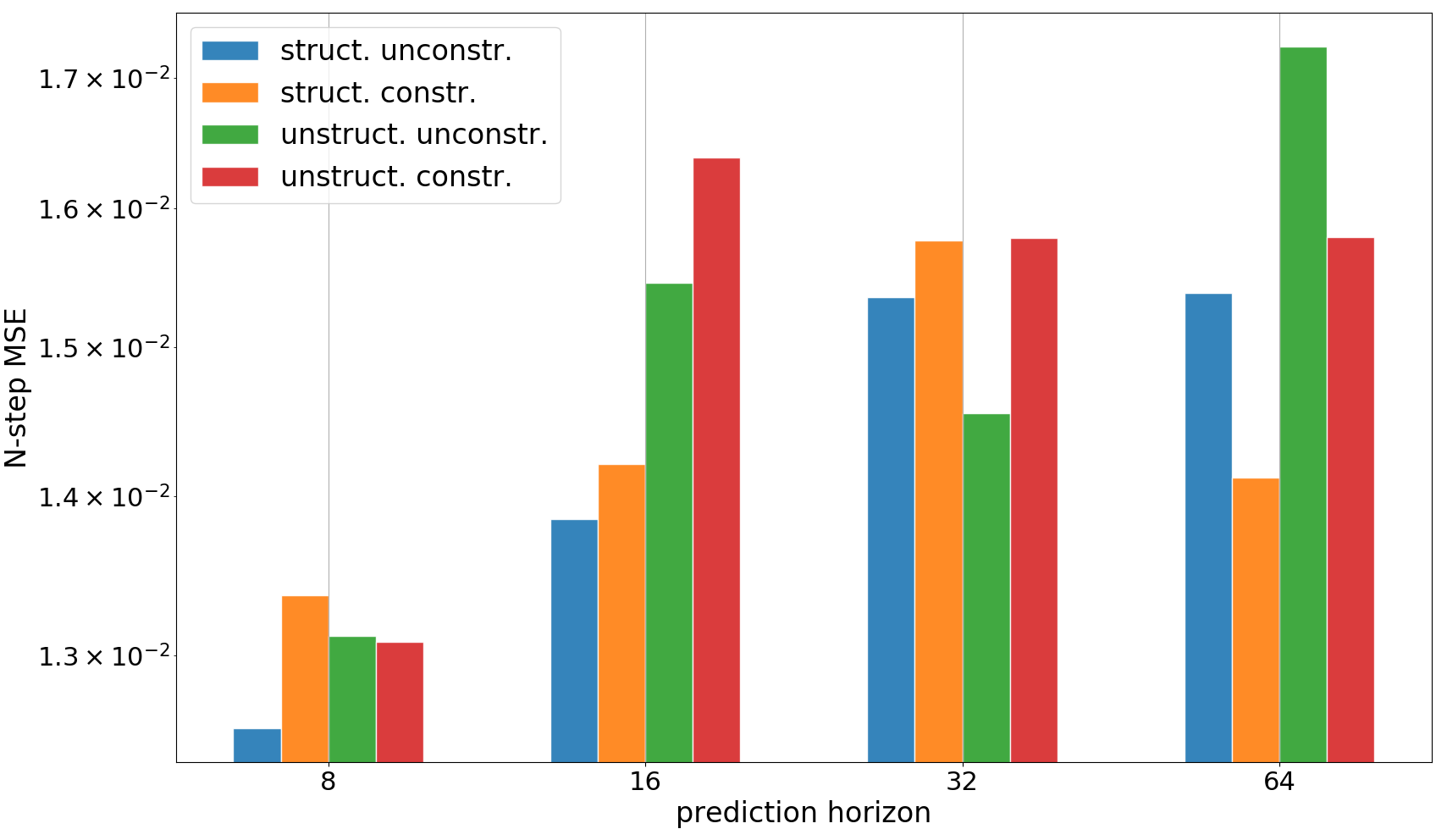}
         \caption{ $N$-step MSE.}
         \label{fig:nstepMSE}
     \end{subfigure}
\caption{Effect of penalty constraints on open-loop and $N$-step ahead MSE evaluated on a test set using structured and unstructured models, with increasing training prediction horizon $N$. }
\label{fig:MSE}
\end{figure*}

\paragraph{Effect of prediction horizon and  penalty constraints}
Fig.~\ref{fig:MSE} shows test set performance with open-loop MSE and $N$-step ahead MSE losses for structured and constrained model variants trained with increasing prediction horizon $N$. 
As expected, Fig.~\ref{fig:nstepMSE} shows that $N$-step MSE rises with a longer prediction horizon in the training loss function because learning long-term predictions is generally a more difficult task. 
Tab.~\ref{tab:best_results} reports larger MSE gaps between $N$-step and open-loop loss for smaller prediction horizon $N=16$. On the other hand, larger horizon $N=64$ minimizes the gap between $N$-step loss function and open-loop performance, hence providing a more accurate assessment of the desired performance measure.
Also, as shown in Fig.~\ref{fig:openMSE},
longer prediction horizon tends to improve the overall open-loop simulation performance of all constrained models.
The same does not hold for unconstrained models for which the performance starts to deteriorate with a horizon longer than $16$. This indicates that including penalty constraints in the training loss function helps to improve the model accuracy over longer prediction horizons.  The intuition here is simple; by confining the system outputs into a physically meaningful subspace, the model is less likely to learn diverging long-term trajectories.

\paragraph{Effect of physics-inspired structure}
Fig.~\ref{fig:MSE} demonstrates that adding building-physics inspired structure into the neural state-space model undeniably improves both open-loop and $N$-step MSE.
Results in Tab.~\ref{tab:best_results} confirm that 
both constraints and structure have a positive influence on the open-loop performance of trained models, 
while structure being a more significant modeling assumption.
Applying both structure and constraints yields a $15\%$ reduction in error against unstructured and unconstrained neural state-space model counterparts.
By decoupling the state, control action, and disturbance dynamics into separate blocks modeled by neural networks, we prevent the model from learning lumped dynamics behavior. Each block can now learn different nonlinear transformations, which can be independently interpreted as structural heat transfer dynamics for states $f_x$, HVAC dynamics for inputs $f_u$, and weather and occupancy thermal dynamics for disturbance signals $f_d$.

\begin{figure*}[!htbp]
\centering
 \begin{subfigure}[b]{0.49\textwidth}
         \centering
        \includegraphics[width=1.0 \textwidth]{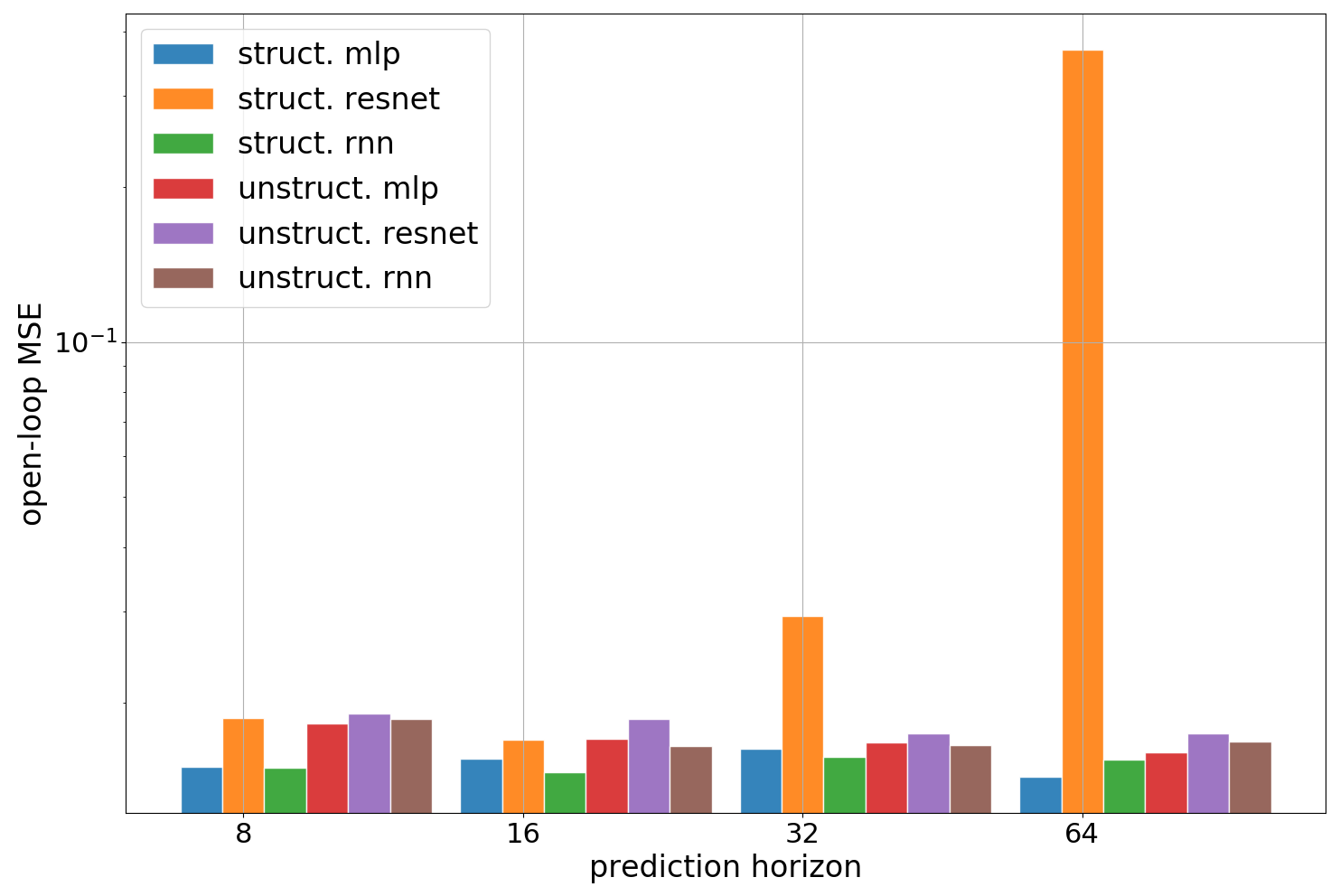}
         \caption{Open-loop MSE.}
         \label{fig:openMSE_nlin}
     \end{subfigure}
     \hfill
     \begin{subfigure}[b]{0.49\textwidth}
         \centering
         \includegraphics[width=1.0 \textwidth]{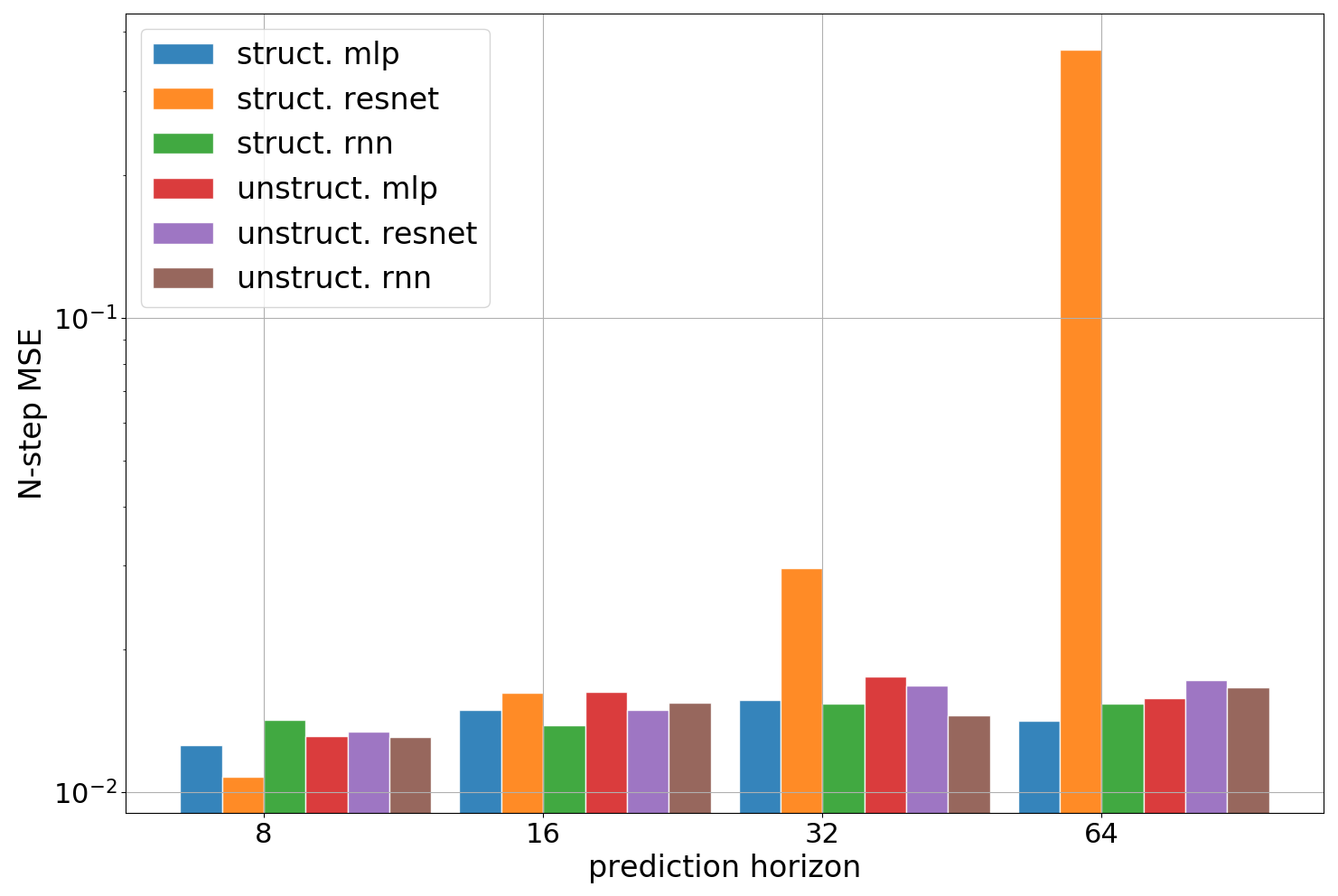}
         \caption{ $N$-step MSE.}
         \label{fig:nstepMSE_nlin}
     \end{subfigure}
\caption{Effect of neural blocks architecture on open-loop and $N$-step ahead MSE evaluated on a test set using structured and unstructured models, with increasing training prediction horizon $N$. }
\label{fig:MSE_nlin}
\end{figure*}

\begin{figure*}[!htbp]
\centering
 \begin{subfigure}[b]{0.49\textwidth}
         \centering
        \includegraphics[width=1.0 \textwidth]{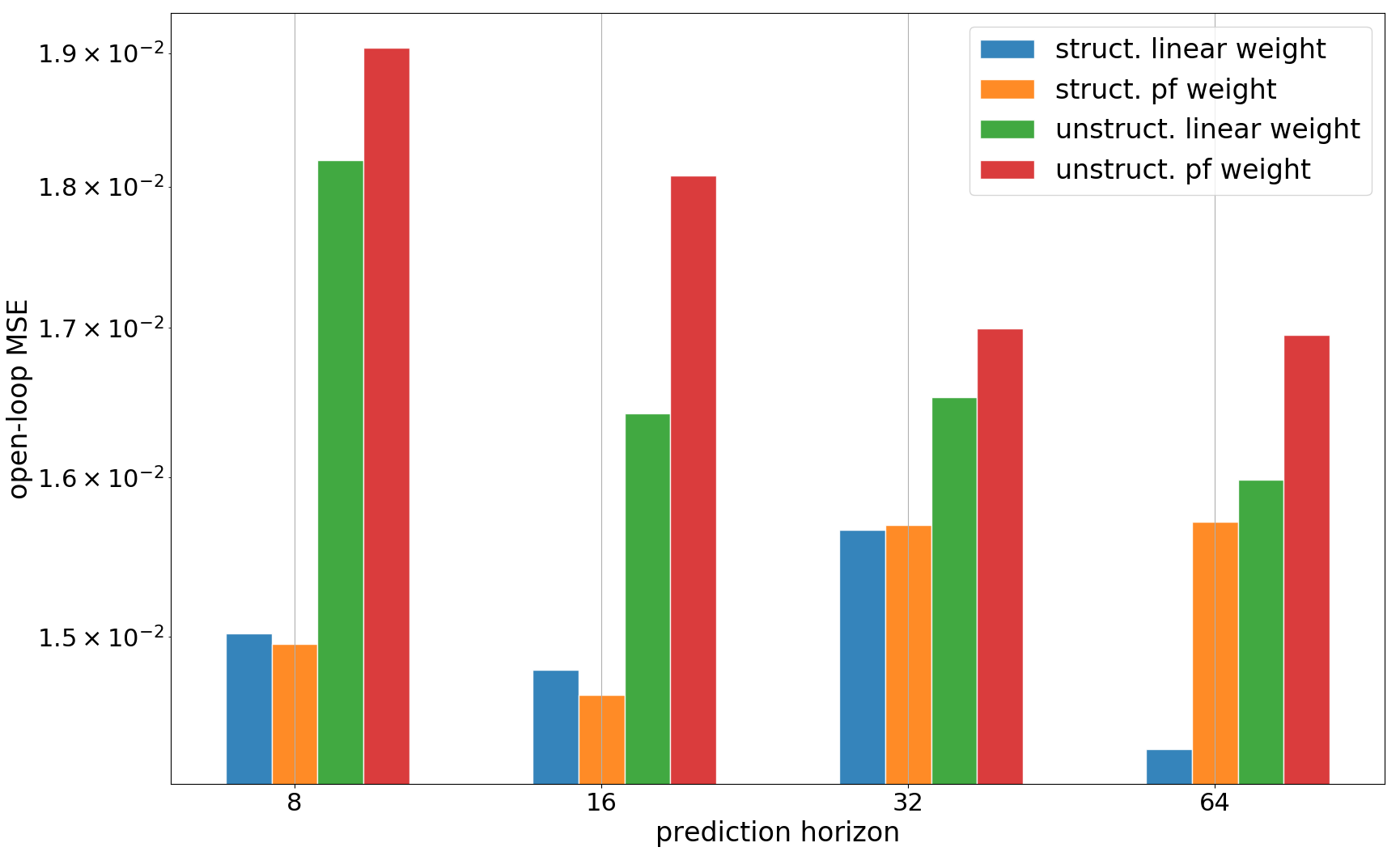}
         \caption{Open-loop MSE.}
         \label{fig:openMSE_lin}
     \end{subfigure}
     \hfill
     \begin{subfigure}[b]{0.49\textwidth}
         \centering
         \includegraphics[width=1.0 \textwidth]{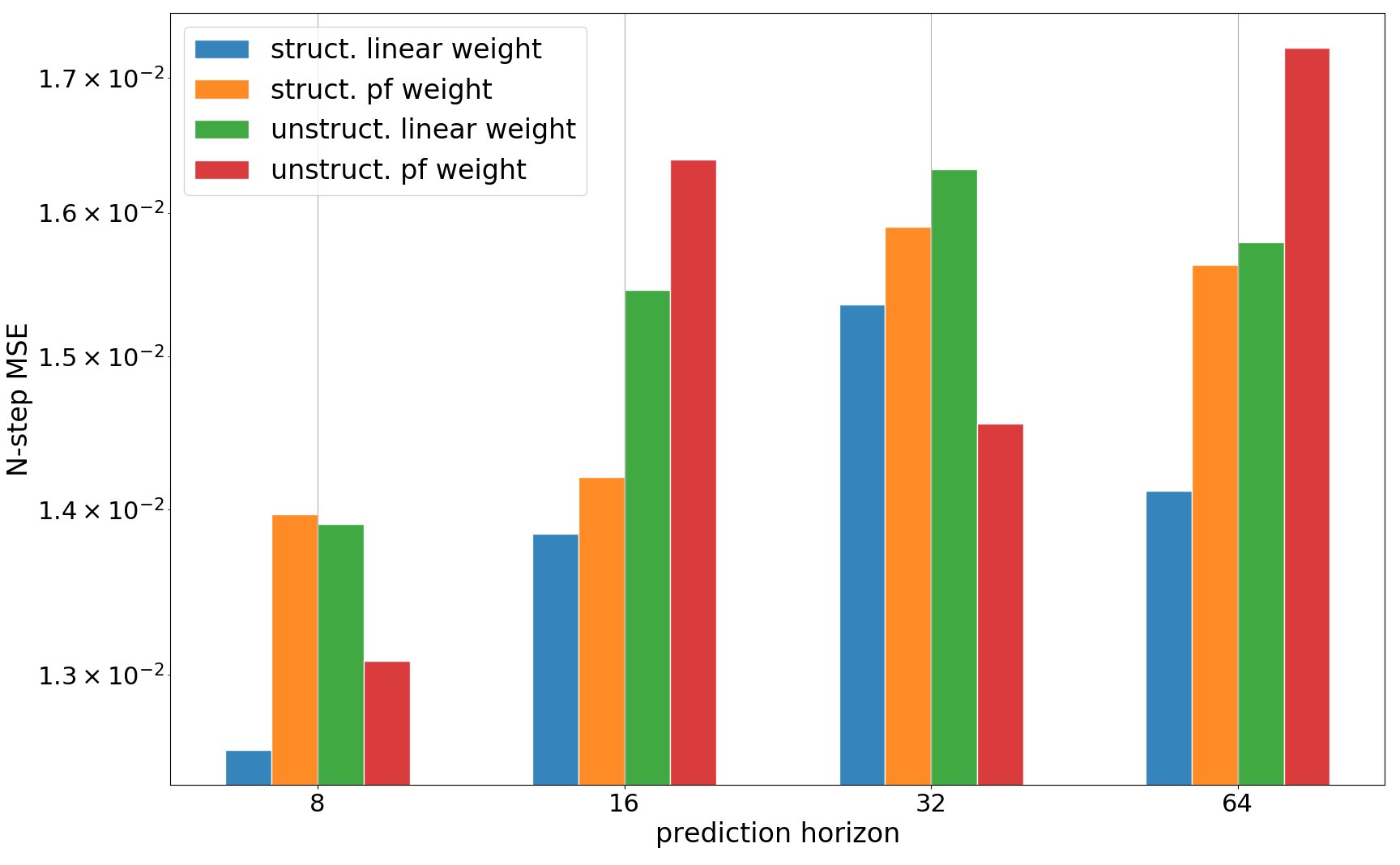}
         \caption{ $N$-step MSE.}
         \label{fig:nstepMSE_lin}
     \end{subfigure}
\caption{Effect of eigenvalue constraints via \texttt{pf} factorization on open-loop and $N$-step ahead MSE evaluated on a test set using structured and unstructured models, with increasing training prediction horizon $N$. }
\label{fig:MSE_lin}
\end{figure*}

\paragraph{Effect of neural blocks architecture}
Fig.~\ref{fig:MSE_nlin} shows the effect on open-loop and $N$-step MSE of using different neural architectures for representing the individual blocks of structured and unstructured neural state-space models, respectively. 
We focus our analysis on best-case open-loop performance displayed in Fig.~\ref{fig:openMSE_lin}. Please note the y-axis is in the logarithmic scale. 
Surprisingly, models with ResNet architectures are less accurate than best performing RNN or MLP across all prediction horizons and deteriorate fast with increasing prediction horizon. The cause of ResNets' poor performance is hard to estimate at this point, and more in-depth analysis needs to be performed in the future.
On the other hand, the performance of models with both RNN and MLP blocks is comparable and scales well also with larger horizons.
While models with RNN blocks tend to perform better for shorter horizons, models with MLP architecture score better for the largest time horizon of $64$ steps.
This might be linked with well known RNN issues, such as vanishing, and exploding gradient problems causing difficulties when learning long-term dependencies~\cite{Pascanu2013,Kolen5264952}.

\paragraph{Effect of weight's eigenvalue constraints}
Fig.~\ref{fig:MSE_lin} shows test set performance with open-loop MSE and $N$-step ahead MSE losses for structured and unstructured model variants with and without eigenvalue constraints via \texttt{pf} factorization of weights.
Due to restrictive nature of the \texttt{pf} factorization, in Fig.~\ref{fig:nstepMSE_lin} we observe larger increase in $N$-step MSE compared to unconstrained  \texttt{linear} weights for most of the cases. However, as shown in 
in Fig.~\ref{fig:openMSE_lin}, the eigenvalue constraints improve the performance of the structured models for shorter prediction horizons, as a consequence of the imposed inductive bias towards learning dissipative heat transfer dynamics. 
On the other hand, unstructured models do not benefit from using \texttt{pf} factorization at all. The reason is that imposed eigenvalue constraints are inspired by the building envelope dynamics exclusively modeled with $f_x$ map of structured models~\eqref{eq:RNN}. In contrast, unstructured models~\eqref{eq:black_box} learn lumped envelope, HVAC, and disturbance dynamics, hence they fail to benefit from any block-specific priors.

\subsection{Eigenvalue Analysis and Physical Interpretability} 
\begin{figure*}[!htbp]
\centering
 \begin{subfigure}[b]{0.31\textwidth}
         \centering
        \includegraphics[width=1.0 \textwidth]{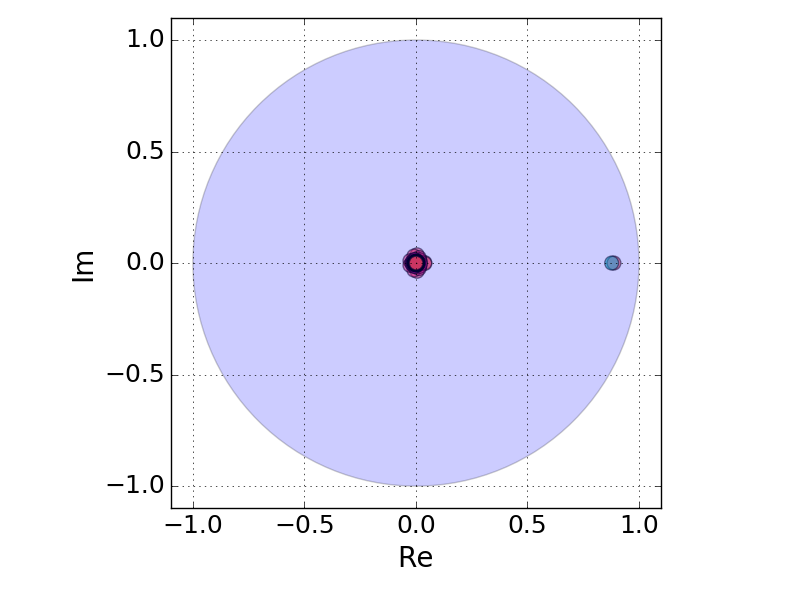}
         \caption{Eigenvalues of \texttt{pf} factorized weights of structured model~\eqref{eq:RNN} dynamics $f_x$.}
         \label{fig:eigen_block_pf}
     \end{subfigure}
     \hfill
     \begin{subfigure}[b]{0.31\textwidth}
         \centering
         \includegraphics[width=1.0 \textwidth]{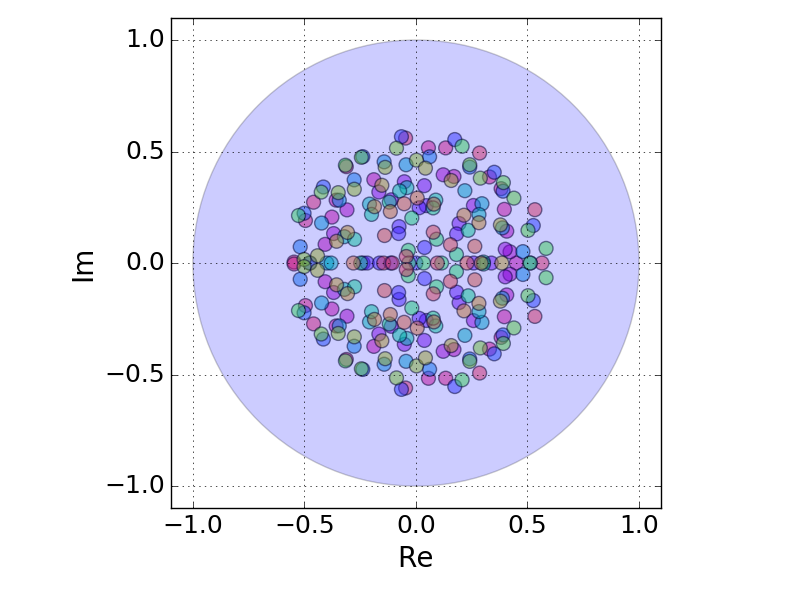}
          \caption{Eigenvalues of unconstrained weights of structured model~\eqref{eq:RNN} dynamics $f_x$.}
         \label{fig:eigen_block_lin}
     \end{subfigure}
          \hfill
     \begin{subfigure}[b]{0.31\textwidth}
         \centering
         \includegraphics[width=1.0 \textwidth]{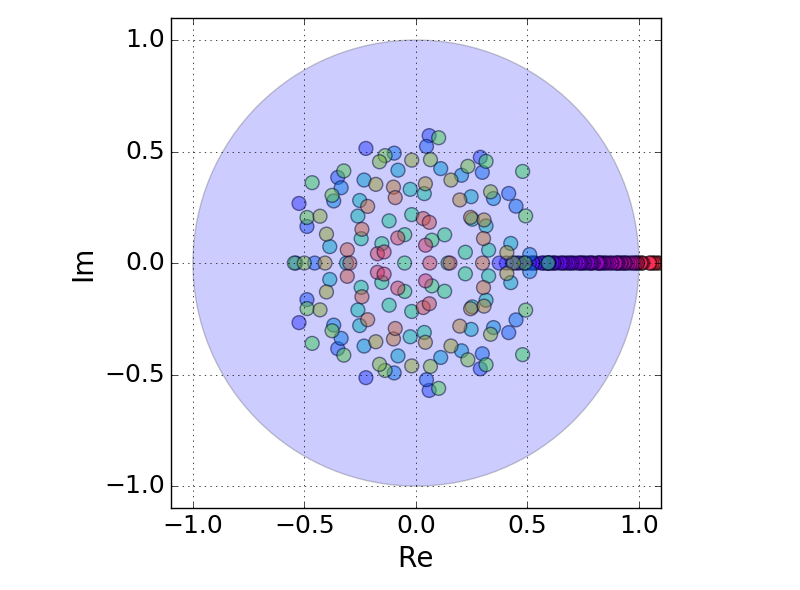}
        \caption{Eigenvalues of unconstrained weights of unstructured model~\eqref{eq:black_box} dynamics $f$.}
         \label{fig:eigen_unstruct_lin}
     \end{subfigure}
\caption{Eigenvalue plots of the weights of system dynamics maps $f_x$, and $f$ of learned structured and unstructured dynamical models, respectively.  Blue circles represent stable regions. }
\label{fig:eigenvalues}
\end{figure*}

 Fig. \ref{fig:eigenvalues} shows concatenated eigenvalues in the complex plane for weights of the state transition maps $f_x$ and $f$ of learned structured~\eqref{eq:RNN} and 
 unstructured~\eqref{eq:black_box} recurrent neural dynamics models, respectively.
 Besides structure Fig. \ref{fig:eigenvalues} compares the effect of  eigenvalue constraints using Perron-Frobenius (\texttt{pf}) factorization
 of the system dynamics weights.
 Please note that we plot only eigenvalues of the neural network's weights. Hence the dynamic effects of the activation functions are omitted in this analysis. 
 However, all our neural network blocks are designed with \texttt{GELU} activation functions, which represent contractive maps with strictly stable eigenvalues.
 Therefore, based on the argument of the composition of stable functions, the global stability of the learned dynamics is not compromised.
 
  Fig.~\ref{fig:eigen_block_pf} shows the effect of proposed eigenvalue constraints \texttt{pf} factorization, and verifies that the dominant eigenvalue remains within prescribed bounds
 $\lambda_{\text{min}} =0.8$ and $\lambda_{\text{max}}=1.0$. Hence the disipativeness of the learned dynamics is hard constrained within physically realistic values when using \texttt{pf} factorization.
Another interesting observation is that there are only two dominant dynamical modes with eigenvalues larger than $0.8$, one per each layer of $f_x$. While the rest of the eigenvalues fall within $0.05$ radius, hence representing less significant dynamic modes~\cite{Schmid2008DynamicMD,DMC2014}.
This indicates a possibility to obtain lower-order representations of the underlying higher-order nonlinear system, a property useful for real-time optimal control applications.

In contrast, as displayed in Fig.~\ref{fig:eigen_block_pf}
and Fig.~\ref{fig:eigen_unstruct_lin},
the eigenvalues of standard unconstrained weights
for both structured  and unstructured models are more dispersed with larger imaginary parts.
The imaginary parts indicate oscillatory modes of the autonomous state dynamics $f_x$ and $f$, respectively. However, in the case of building thermal dynamics, the periodicity of the dynamics is caused by external factors such as weather and occupancy schedules. From this perspective, the structured models using \texttt{pf} factorization of the weights, are closer to the physically realistic parameterization of the system dynamics.
Additionally, not using eigenvalue constraints may result in learning unstable weights. Fig.~\ref{fig:eigen_unstruct_lin} displays an example where the unstructured learned model does not guarantee the satisfaction of physically realistic dissipativeness property.

\section{Conclusions} \label{sec:clc}

Control-oriented system dynamic models are indispensable parts of most advanced building control strategies, such as model predictive control.
Consequently, reliable data-driven modeling methods that are cost-effective in terms of computational demands, data collection, and domain expertise have the potential to revolutionize the field of energy-efficient building operations through the wide-scale acquisition of building specific, scalable, and accurate prediction models.
Here we presented a constrained deep learning method for sample-efficient and physics-consistent data-driven modeling of building thermal dynamics. Our approach does not require the large time investments by domain experts and extensive computational resources demanded by physics-based emulator models. Based on only $10$ days' measurements, we significantly improve on prior state-of-the-art results for a modeling task using a real-world large scale office building dataset. 
This improvement in model predictions is attributed to our contributions of structural assumptions, eigenvalue constraints imposed on weights, and penalty methods imposed on the outputs of a deep recurrent neural network model.
We systematically analyze the added value of
the proposed structural assumptions and constraints in comparison with their unstructured and unconstrained model counterparts.
Additionally, we assess the interpretability and level of physical realism of the learned system dynamics by connecting the neural weights' eigenvalue analysis with known dynamical properties of buildings.
The modeling results using a real-world dataset obtained from a large-scale office building demonstrate the accuracy, data-efficiency, and interpretability of the proposed method.
A potential limitation of the presented approach is the restrictiveness of the used constraints, where wrong initial guess of the eigenvalue and penalty constraints bounds may lead to decreased accuracy of the learned model. Hence, some level of engineering insight is required to properly use the presented methodology for modeling building's thermal dynamics.
Future work includes a systematic comparison against physics-based emulator models and other standard data-driven methods.
Authors also plan to use the method as part of advanced predictive control strategies for energy-efficient operations in real-world buildings.

\section*{ACKNOWLEDGEMENT} \label{sec:ack}
This work was funded by the Physics Informed Machine Learning (PIML) investment at the Pacific Northwest National Laboratory (PNNL). 
This work emerged from the IBPSA Project 1, an international project conducted under the umbrella of the International Building Performance Simulation Association (IBPSA).

\bibliographystyle{unsrt}
\bibliography{bib}

\end{document}